\documentclass{ifacconf}
\usepackage{tabularx}
\usepackage{booktabs}
\usepackage{graphicx}      % include this line if your document contains figures
\usepackage[numbers]{natbib}        % required for bibliography
\usepackage{amsmath}
\usepackage{amsfonts}

\begin{document}
\begin{frontmatter}

\title{Establishing a Dynamic Multimodal HRI Dataset for Engagement Analysis with a Humanoid Robot\thanksref{footnoteinfo}} 

\thanks[footnoteinfo]{This research was supported by Incheon National University Research Grant in 2025 (\#2025-0391).} 

\author{Buwan Kim$^1$, Wonse Jo$^1$}

\address{$^1$Department of Information and Telecommunication Engineering,}
\address{Incheon National University, Incheon, South Korea 22012. \\
\{boowan,jow\}@inu.ac.kr}

\begin{abstract}                % Abstract of 50--100 words   
This paper presents an experimental design for constructing a multimodal dataset to analyze user engagement in human–robot interaction (HRI). Prior studies have mainly relied on observable behavioral cues, with limited frameworks integrating physiological signals. We therefore propose a structured data-collection protocol to build a multimodal dataset that includes wearable physiological signals, behavioral data, and self-report measures under different levels of task complexity defined in this experiment.
%The contribution lies in the experimental design and multimodal dataset configuration.
%%% Stat + DL classification accuracy --> validation of dataset quality

%Furthermore, we demonstrate the utility of this dataset using a Transformer-based architecture to validate the temporal correlations between these heterogeneous modalities. 

\end{abstract}

\begin{keyword}
Human–Robot Interaction, Multimodal Dataset,Physiological Signals, Kinematic Synchronization, Experimental Design
\end{keyword}

\end{frontmatter}
%===============================================================================

\section{Introduction}
With the advances in automation technologies, service robots are increasingly deployed in industrial and professional environments. Beyond simple information delivery, robots are expected to engage in task-oriented human–robot collaboration in domains such as manufacturing and logistics \citep{Villani2018}. In these contexts, collecting multimodal interaction data is essential for studying engagement during collaborative activities.

Existing HRI research has examined engagement primarily through observable behavioral cues and, in some cases, through additional sensing modalities \citep{Sorrentino2021}. Behavioral indicators such as gaze, facial expressions, and posture have been widely used to analyze interaction states \citep{Oertel2014}. Physiological signals have also been incorporated in certain datasets to capture internal responses during interaction \citep{Anzalone2015}.

Building on these research directions, this paper proposes a conceptual design of a user experiment to build a multimodal HRI dataset that integrates behavioral cues, physiological signals, and synchronized motion measurements from both the user and the robot. The proposed scenario in this user experiment includes collaborative tasks such as object sorting and transport under varying levels of task complexity. The primary objective is to define a structured data acquisition protocol and sensing configuration that combine these modalities within a unified experimental setting.

\begin{table*}[t] 
\centering
\caption{Key Feature Comparison between Proposed and Existing Multimodal Datasets}
\label{tb:comparison_wide}
\small
\renewcommand{\arraystretch}{0.9}

\begin{tabular}{@{}lllllll@{}}
\toprule
\textbf{Paper (year)} & \textbf{Platform} & \textbf{Scenario} & \textbf{VISION} & \textbf{Physio.} & \textbf{IMU} \\ \midrule

UE-HRI (2017) 
& Wheel-Based Humanoids (Pepper) 
& Simple Move 
& O (Face, Gaze) 
& X 
& X \\ \midrule

Kompatsiari et al. (2019) 
& Leg-Based Humanoids (iCub) 
& Social Gaze 
& O (Face, Gaze) 
& X 
& X \\ \midrule

MHHRI (2015) 
& Leg-Based Humanoids (Nao) 
& Social Dialogue 
& O (Face, Body) 
& O (EDA, ECG) 
& X \\ \midrule

Confusion Detection (2023) 
& Wheel-Based Humanoids (Pepper) 
& Language Task 
& O (Exp., Gaze) 
& X 
& X \\ \midrule

Ours (Proposed) 
& Leg-Based Humanoids (Unitree U1) 
& \begin{tabular}[c]{@{}l@{}}Interactive Task with\\ 3-Level Complexity\end{tabular} 
& O (Dist., Pos.) 
& O (EDA, PPG) 
& O \\ 

\bottomrule
\end{tabular}
\end{table*}

\section{Related Work}

Based on existing literature, prior HRI studies analyzed engagement using observable behavioral cues such as gaze, facial expressions, and posture \citep{Abrougui2017, Kompatsiari2018, Vanneste2021}. The UE-HRI dataset focuses on spontaneous interaction using vision-based behavioral indicators \citep{Abrougui2017}, while gaze-based engagement analysis was explored in controlled settings \citep{Kompatsiari2018}. The HRI-Confusion dataset investigates user confusion during interaction \citep{Vanneste2021}. Engagement modeling and assessment approaches were explored in spontaneous HRI scenarios \citep{BenYoussef2019}.

In addition, certain multimodal datasets incorporate physiological signals to monitor internal responses during interaction \citep{Anzalone2015, Oertel2014, Sorrentino2021}. The MHHRI dataset includes electrodermal activity (EDA) alongside audiovisual recordings \citep{Anzalone2015}. Systematic analyses further emphasized that multimodal approaches that integrate behavioral, physiological, and subjective indicators improve the robustness and reliability of engagement assessment in HRI contexts \citep{Oertel2014, Sorrentino2021}. These works extend engagement analysis beyond observable behavior by incorporating autonomic measurements and highlighting the complementary role of subjective self-report measures \citep{Betella2016, Sorrentino2021}.

Building upon prior works highlighting the importance of structured interaction in human--robot collaboration \citep{Nikolaidis2013}, this study proposes a dataset design that integrates behavioral indicators, physiological signals including EDA and photoplethysmography (PPG), and synchronized Inertial Measurement Unit (IMU) measurements collected from both the user and the robot within a collaborative task scenario.

%Based on our literature review, although there were some research works exploring engagement measures and enhancing engagement in human-robot interaction, there still exists a literature gap about how to enhance the engagement level in real-time based on a comprehensive assessment of the visitors' preference and engagement with a tour-guide robot.

% 3장 시작 직후 실험 절차도를 배치 (파일명은 예시로 Procedure.pdf라 명명)
\begin{figure}[t]
    \centering
    \includegraphics[width=1\linewidth]{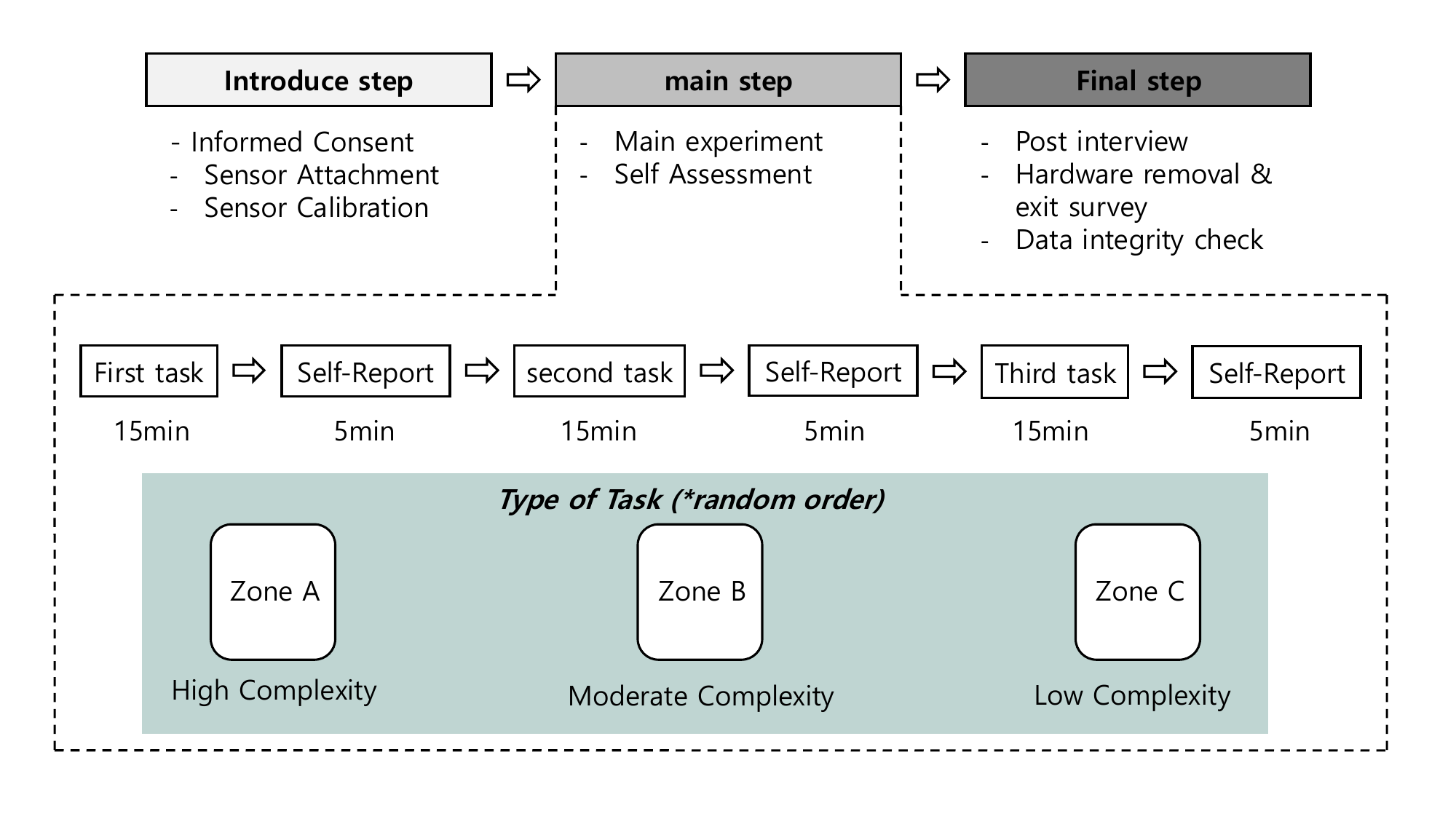} 
    \vspace{-15pt}
    \caption{Experimental procedure structured into introduction, main tasks across randomized zones, and final data integrity check.}
    \label{fig:experimental_procedure}
\end{figure}

\section{Proposed Experimental Protocol}
The experimental session plans to follow a structured protocol as illustrated in Figure~\ref{fig:experimental_procedure}. The process is designed to capture stable baseline data and subsequent behavioral shifts during collaborative tasks.

\begin{figure*}[t]
  \centering
  \includegraphics[width=0.95\textwidth]{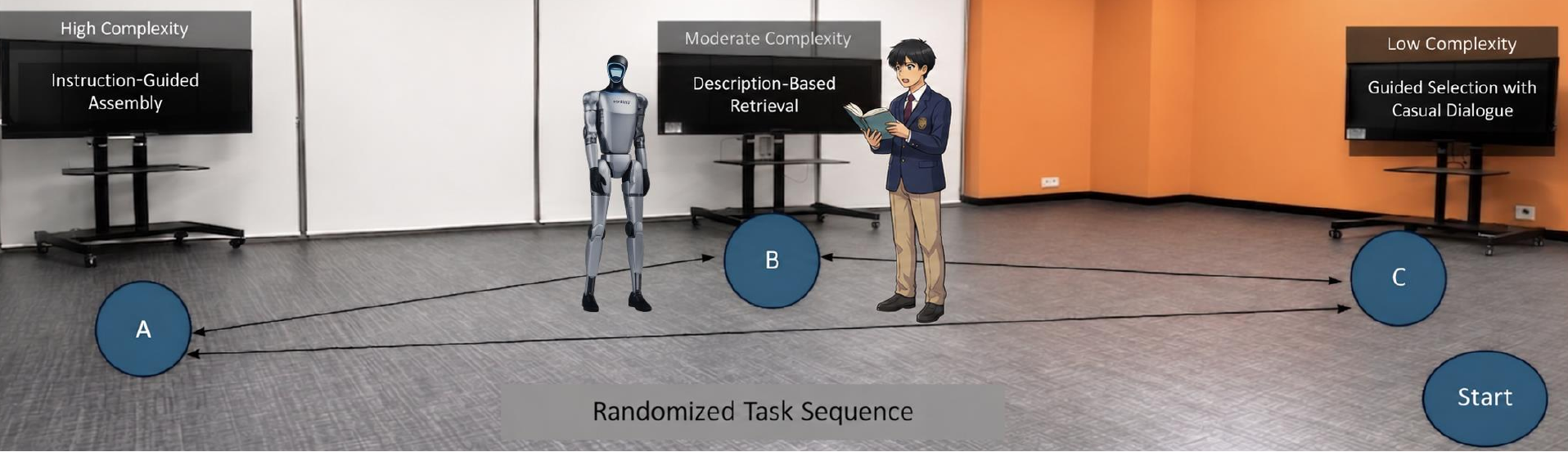}
  \caption{Overview of the system structure.}
  \label{fig:experimental_setup_wide}
\end{figure*}

\subsection{Scenario: Multi-Stage Collaborative Tour Task}
The experiment is planned to be conducted in a controlled laboratory environment to ensure reliable signal acquisition. After informed consent, the session begins with a brief introductory interaction, followed by collaborative tasks with the robot.

A within-subject design is employed, in which each participant completes three conditions with different complexity levels. The order of the task zones is randomized to mitigate potential order effects. To simulate a realistic tour-guide scenario, the protocol integrates short-range coordinated movement with collaborative object-handling tasks. Lightweight secondary-task elements are incorporated in some conditions to introduce variations in attentional allocation and engagement during interaction. Detailed interaction structures for each condition are described in Section 3.3.

\subsection{Participants and Inclusion Criteria}

A total of 30 participants will be recruited through university mailing lists and online bulletin board postings. Participation will be voluntary, and no prior experience with robotics is required. The study will be conducted in a controlled laboratory environment, and participants will attend the session in person. Participants will receive a small monetary compensation for their participation.

For this experiment, the sample size is determined through an a priori power analysis conducted using G*Power \citep{faul2007g}, assuming an effect size of \textit{f} = 0.24, a significance level ($\alpha$) of .05, and a statistical power of .80.

To ensure the reliability and validity of the multimodal physiological measurements, the following inclusion and exclusion criteria are applied:
\begin{itemize}
    \item \textbf{Inclusion Criteria:} Participants must have 7 or corrected-to-normal vision and be physically capable of performing collaborative tasks with a humanoid robot.
    \item \textbf{Exclusion Criteria:} Individuals with diagnosed cardiovascular or neurological disorders, or those taking medications known to significantly affect autonomic nervous system activity (e.g., heart rate or skin conductance), will be excluded.
\end{itemize}

All participants will receive a full explanation of the experimental procedure and will provide written informed consent prior to beginning the experiment.

\subsection{Experimental Design}

The collaborative task will consist of three levels of complexity as follows:

\textbf{Zone A (High Complexity):}
Participants will retrieve five specified components by following the robot’s instructions. 
The robot will provide real-time validation and corrective feedback during execution. 
After retrieval, the components will be aligned in a predefined sequence, resulting in frequent instruction--response exchanges.

\textbf{Zone B (Moderate Complexity):}
In this condition, the robot will provide property-based descriptions instead of explicitly naming the target items. 
Participants will interpret the descriptions and complete the required sequence independently. 
Feedback will be delivered only after the participant indicates task completion.

\textbf{Zone C (Low Complexity):}
The task will be performed alongside brief casual dialogue, beginning with a short greeting phase. 
Participants will sequentially select and align five clearly specified components. 
Immediate confirmation will be provided after each action.

Across all zones, multimodal signals will be recorded in a synchronized manner. 
These will include physiological data (EDA, PPG), wearable IMU data (EmotiBit), robot-side IMU measurements, 
joint encoder data, 3D LiDAR, and RGB-D streams.

After completing each zone, participants will complete a 15-item questionnaire rated on a 5-point Likert scale, and the responses will be stored together with the corresponding task condition.

The system is organized into two synchronized streams. The Robot-centric stream captures robot kinematics and environmental context through joint encoders, IMU, 3D LiDAR, and RGB-D data. The Human-centric stream records physiological and motion-related measurements from the participant, including EDA, PPG, and wearable IMU signals.

The objective of this study is to propose a multimodal data architecture that synchronizes heterogeneous signals collected during human–robot collaboration.

\section{Multimodal Data Architecture}

\subsection{Subjective Measures}
% Survey --> Survey items (questions) 

The validated 15-item questionnaire will be used for post-task subjective evaluation \cite{BenYoussef2019} and included in the dataset as ground-truth data. The expected questionnaire is summarized in Table~\ref{tab:questionnaire}. All items are rated on a 5-point Likert scale.

\begin{table}[ht]
\centering
\caption{Subjective evaluation metrics for HRI quality \cite{BenYoussef2019}}
\label{tab:questionnaire}
\footnotesize % 한 단에 넣기 위해 폰트 크기를 더 줄임
\renewcommand{\arraystretch}{1.1} % 줄 간격을 약간 조절
\tabcolsep=3pt % 열 사이 간격을 좁혀서 공간 확보
\begin{tabularx}{\columnwidth}{l l X} % \textwidth 대신 \columnwidth 사용
\toprule
\textbf{ID} & \textbf{Category} & \textbf{Questionnaire Item} \\ \midrule
Q1 & Cog. Load & I felt that the task was mentally demanding and complex. \\
Q2 & Comp. & The robot understood my verbal instructions accurately. \\
Q3 & Relev. & The robot’s responses were appropriate and relevant. \\
Q4 & Ease & It was effortless to communicate with the robot. \\
Q5 & Clarity & The robot's voice was clear and easy to comprehend. \\
Q6 & Nat. & The robot's movements were natural and lifelike. \\
Q7 & Engag. & I was fully concentrated in the collaboration. \\
Q8 & Attent. & The robot was consistently paying attention to me. \\
Q9 & Social & The robot felt like a real social companion. \\
Q10 & Latency & The robot's response time was appropriate. \\
Q11 & Robust. & There were no technical errors or interruptions. \\
Q12 & Utility & The info provided by the robot was useful. \\
Q13 & Retent. & I have a strong intention to interact again. \\
Q14 & Incl. & I did not feel ignored or neglected by the robot. \\
Q15 & Satisf. & Overall, I am highly satisfied with the interaction. \\ \bottomrule
\end{tabularx}
\end{table}

Subjective interaction quality is assessed using a 15-item post-task questionnaire administered after each task zone. All items are rated on a 5-point Likert scale. The questionnaire captures perceived cognitive demand, communication quality, responsiveness, social perception, and overall satisfaction. These self-reported responses are included in the dataset as subjective annotations corresponding to each task condition.

\subsection{Objective Measures}
% biosensors : sensor types 
% camera data: from robot
% Robot sensor data
There are two types of data streams for collecting objective and subjective data for this multimodal dataset:

\subsubsection{Human-Centric Stream:}
Physiological signals, including EDA and PPG, are collected using a wearable EmotiBit device \citep{Montgomery2023EmotiBit}. Wrist-level IMU data are simultaneously recorded to capture participant motion dynamics.

\subsubsection{Robot-Centric Stream:}
Joint encoder readings and robot-side IMU measurements are recorded to represent articulated motion and body dynamics of the robot platform. In addition, RGB camera streams and depth-based spatial measurements are logged to provide contextual information regarding relative positioning and interaction environment.

All objective data streams are temporally synchronized within a unified logging framework through the Robot Operating System 2 (ROS2) \citep{Macenski2022}, ensuring consistent multimodal alignment and enabling real-time data collection.

\section{Data Synchronization and Preparation}

\begin{figure}[htbp]
    \centering
    \includegraphics[width=1\linewidth]{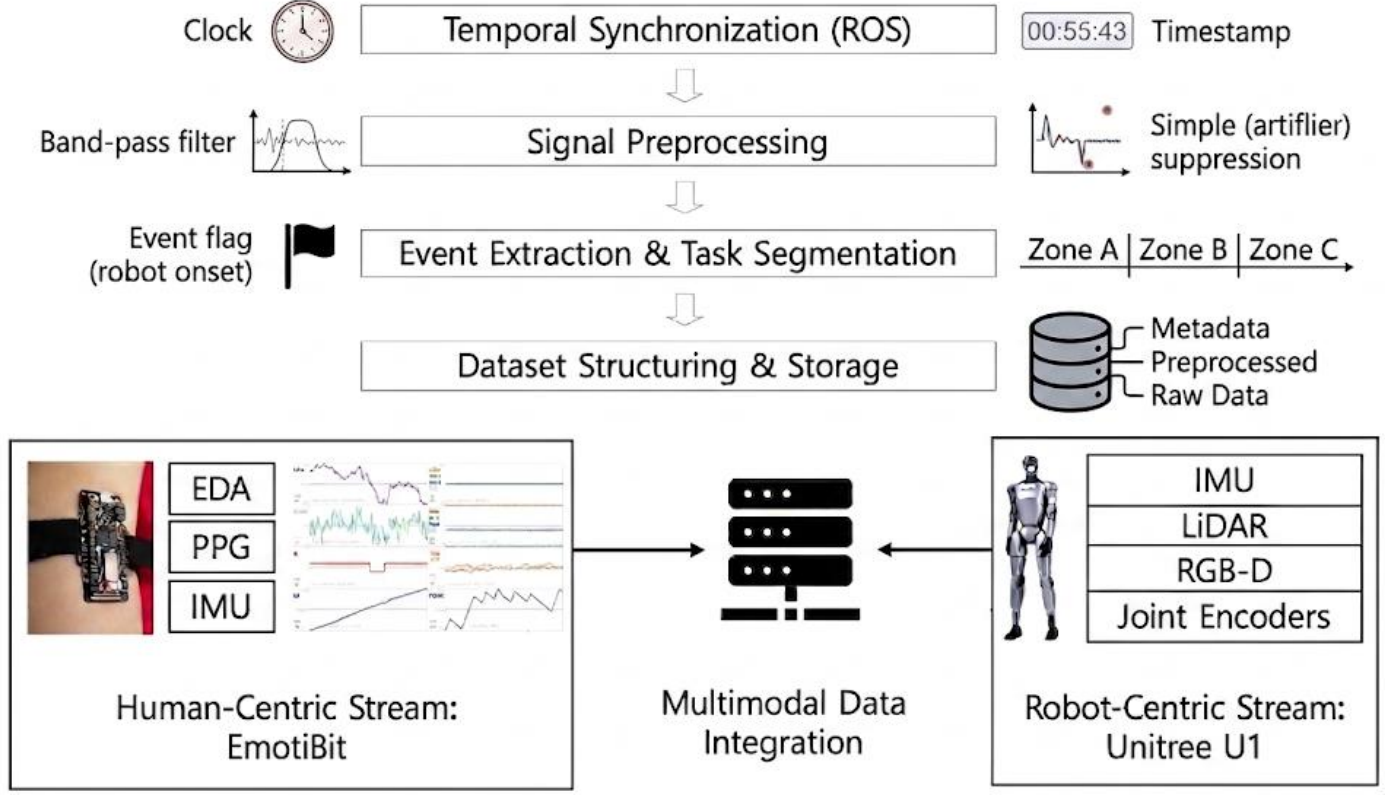} 
    \vspace{-15pt}
    \caption{Overview of the multimodal data processing framework.}
    \label{fig:experimental_procedure}
\end{figure}

To ensure consistency across heterogeneous data streams, all signals are planned to be timestamped using a unified acquisition clock and temporally aligned via resampling onto a common timeline.

Physiological signals (EDA and PPG) are designed to undergo band-limited filtering to suppress low- and high-frequency noise, with outlier removal to mitigate motion-related artifacts. Both raw and filtered signals are intended to be retained to preserve analytical flexibility.

Human- and robot-side IMU data are planned to be processed through gravity compensation, coordinate normalization, and unit standardization to ensure comparability across motion streams. Joint encoder readings are designed to be converted into angular velocity profiles, and velocity-based thresholding is intended to extract robot motion onset timestamps for structured segmentation.

RGB and depth-based spatial measurements are planned to be temporally aligned and frame-sampled as needed, with relative distance and positional information stored as scalars to support multimodal analysis.

Task-zone segmentation (Zone A/B/C) is designed to be performed using protocol timestamps, and robot event logs (e.g., instruction onset and task completion) are intended to serve as auxiliary markers. Both raw and preprocessed datasets are planned to be stored with metadata describing sensor configuration, coordinate conventions, and preprocessing parameters to ensure reproducibility and future usability.

\section{Conclusion}
This paper presents a proposed experimental design for constructing a multimodal dataset in human–robot collaboration scenarios. The framework integrates physiological signals, human-side motion data, robot kinematic measurements, and spatial context streams within a unified, time-synchronized architecture.

By organizing the protocol across three levels of task complexity, the design supports structured comparative analysis of interaction patterns under varying collaborative conditions. In addition to objective sensor streams, subjective post-task assessments are incorporated to provide complementary annotations for each task zone.

The resulting data architecture establishes a foundation for future multimodal analysis in HRI research and is intended to facilitate subsequent investigations into interaction dynamics and user-state modeling within collaborative robotic environments.

\section{Future Work} %%% Eesimtating human status/conditions/ enagement ...
Future work will focus on implementing the proposed protocol and collecting a large-scale multimodal dataset under the described collaborative scenarios. The acquired data will enable systematic investigation of multimodal interaction dynamics across varying task complexity levels.

Subsequent research will explore data-driven modeling approaches to analyze synchronized physiological, kinematic, and spatial streams. The proposed dataset is expected to support future studies on multimodal interaction modeling and adaptive human–robot collaboration.

\footnotesize
\bibliography{ifacconf}      

\end{document}